\documentclass[english]{article} 

\usepackage[T1]{fontenc}
\usepackage[latin9]{inputenc}
\usepackage{babel}
\usepackage[table]{xcolor}
\usepackage{collcell}
\usepackage{hhline}
\usepackage{pgf}
\usepackage{multirow}

\def\colorModel{rgb} 

\newcommand\ColCell[1]{
  \pgfmathparse{#1<50?1:0}  
    \ifnum\pgfmathresult=0\relax\color{white}\fi

\pgfmathsetmacro\compA{100(1-#1/100)}      
\pgfmathsetmacro\compB{100(1-#1/100)} 
\pgfmathsetmacro\compC{100(1-#1/100)}      

  \edef\x{\noexpand\centering\noexpand\cellcolor[\colorModel]{\compA,\compB,\compC}}\x #1
  }
\newcolumntype{E}{>{\centering\arraybackslash\collectcell\ColCell}m{0.65cm}<{\endcollectcell}}  

\usepackage[final]{corl_2017} 

\usepackage{float} 
\usepackage{multirow} 
\usepackage{mathtools} 

\usepackage{amsmath} 
\usepackage{amssymb} 
\usepackage{verbatim} 
\usepackage{array}

\usepackage{siunitx} 
\usepackage{tikz} 
\usepackage{pgfplots}
\pgfplotsset{compat=1.7}

\usepackage{fancyhdr}
\title{%
Driver Distraction Identification with an Ensemble of Convolutional Neural Networks
}

\author{
  Hesham M. Eraqi\\
  Dept. of Computer Science \& Engineering\\
  The American University in Cairo\\
  \texttt{heraqi@aucegypt.edu}\\
  \And
  Yehya Abouelnaga\\
  Dept. of Informatics\\
  Technical University of Munich\\
  \texttt{yehya.abouelnaga@tum.de}\\
  \And
  Mohamed H. Saad \\
  Dept. of Computer and Systems Engineering\\
  Ain Shams University\\
  \texttt{muhamedhusseinsaad@gmail.com}\\
  \And
  Mohamed N. Moustafa\\
  Dept. of Computer Science \& Engineering\\
  The American University in Cairo\\
  \texttt{m.moustafa@aucegypt.edu} \\
}

\begin{document}

\maketitle


\begin{abstract}%
The World Health Organization (WHO) reported 1.25 million deaths yearly due to road traffic accidents worldwide and the number has been continuously increasing over the last few years.
Nearly fifth of these accidents are caused by distracted drivers.
Existing work of distracted driver detection is concerned with a small set of distractions (mostly, cell phone usage).
Unreliable \textit{ad-hoc} methods are often used.
In this paper, we present the first publicly available dataset for driver distraction identification with more distraction postures than existing alternatives.
In addition, we propose a reliable deep learning-based solution that achieves a 90\% accuracy. The system consists of a genetically-weighted ensemble of convolutional neural networks, we show that a weighted ensemble of classifiers using a genetic algorithm yields in a better classification confidence. We also study the effect of different visual elements in distraction detection by means of face and hand localizations, and skin segmentation.
Finally, we present a thinned version of our ensemble that could achieve 84.64\% classification accuracy and operate in a real-time environment.
\end{abstract}



\section{Introduction}

Over the 20 years from 1980 to 2000, the number of licensed drivers in the United States increased by 23.7\%, to reach 190.6 million licenses \cite{liang2009detecting}. From 1990 to 2000, the urban vehicles miles traveled increased by 80\%, while roads building rose by only 37\% \cite{downs2004traffic}; this makes driving a common activity for many people and makes the driving safety an important issue in everyday life. Construction of new roads did not keep up with the noticeable increase in vehicles, leading to more traffic congestion \cite{downs2004traffic}.
In addition, In-Vehicle Information Systems (IVISs) such as media players and navigation devices introduce more distraction to the driving experience and lead to more accidents.

\begin{figure*}[t]
  \centering
  \includegraphics[width=\textwidth]{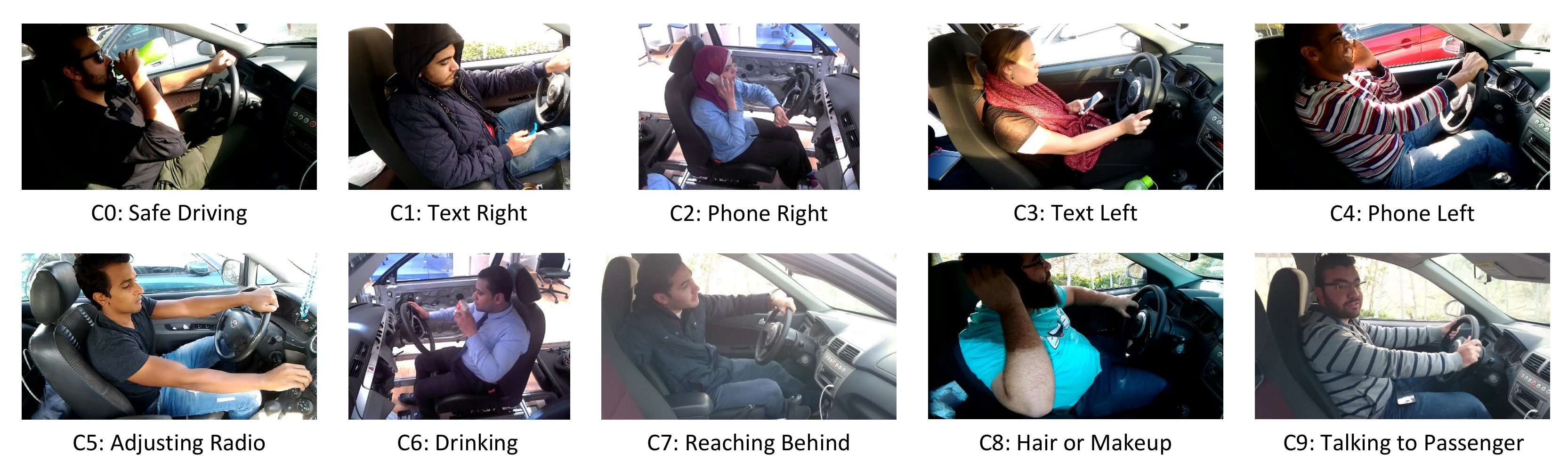}
  \caption{Ten Classes of Driver Postures from our Dataset}
  \label{fig:data_samples}
\end{figure*}

Despite of safety improvements in road and vehicle design, the total number of fatal crashes still increases \cite{liang2009detecting}. The 2017 Global Status Report of the World Health Organization (WHO) reported an estimated 1.25 million yearly deaths due to road traffic accidents worldwide, with up to 50 million people sustaining non-fatal injuries as a result of road traffic accidents \cite{who2017healthstats}. To put that mortality rate into perspective, it is the same as having around seven Boeing airplanes crashing or disappearing every day.
Car accidents mortality rate compares to those of serious diseases (i.e. Hepatitis \& HIV). 
At the same year, the number of global deaths attributable to Hepatitis and HIV are estimated to be in the order of 1.3 million \cite{who2017deathcauses} and 1.1 million \cite{who2017healthstats}, respectively-- which is almost the same as the number of the people dying yearly due to road traffic accidents. Moreover, road traffic accidents cause a huge property damage and the number of road accidents due to distracted driving is steadily increasing.

Nearly fifth of traffic accidents are caused by a distracted driver according to the National Highway Traffic Safety Administration (NHTSA), and approximately 90\% of road accidents being due to human errors in the United States \cite{strickland2013autonomous}. Despite that vehicle crashes are attributed to multiple causes, driver error represents a dominant factor \cite{lee_driving_safety}. In 2015, 3,477 people were killed, and 391,000 were injured in motor vehicle crashes involving distracted drivers \cite{USDepartmentofTrans2017}. The major cause of these accidents was the use of mobile phones \cite{USDepartmentofTrans2017}.

The NHTSA defines distracted driving as "any activity that diverts attention from driving", including: a) talking or texting on one's phone, b) eating and drinking, c) talking to passengers, or d) fiddling with the stereo, entertainment, or navigation system \cite{USDepartmentofTrans2017}.
The Center for Disease Control and Prevention (CDC) provides a broader definition of distracted driving by taking into account visual (i.e. taking one's eyes off the road), manual (i.e. taking one's hands off the driving wheel) and cognitive (i.e. taking one's mind off driving) causes \cite{Services2016}.

One way to address the problem of distracted driver is to develop distraction mitigation systems, which adapt IVIS functions according to driver state. In such a mitigation system, correctly identifying driver distraction is critical, which is the focus of this work. We envision a future where smart vehicles could detect and identify such distraction, and warn the driver against it or take preventive measures. On the other hand, such detection systems can help law enforcement to identify distraction on highway using radar cameras, and penalize certain forms of distraction. Moreover, the recent commercial semi-autonomous cars require drivers to pay attention to the road \cite{eriksson2017takeover}. Autonomous steering control \cite{SteeringControl_2017_NIPS} systems require drivers to be ready to take back control of the wheel \cite{eriksson2017takeover}. This is what makes distracted drivers detection an important system component in these cars. Distraction detection can also be used to enable Advanced Driver Assistance Systems (ADAS) features \cite{FreeSpaceDetection_2017} like Collision Avoidance Systems (CAS) that have to plan evasive maneuvers \cite{CollisionAvoidance_ecta16}.

Research in the field of distracted driving detection follows the definitions presented in \cite{USDepartmentofTrans2017} and \cite{Services2016}.
It detects manual, visual, or cognitive types of distractions.
Cognitive distractions deal with tasks of listening, conversing, daydreaming, or just becoming lost in thought.
In this form of distraction, the driver is ``mentally'' distracted from safe driving even though they are in a safe driving posture.
Visual distractions often refer to situations where the driver takes their eyes off the road due to either ``the presence of salient visual information away from the road causing spontaneous off-road eye glances and momentary rotation of the head'' or the use of multimedia devices (i.e. cell phones, navigation or entertainment systems) \cite{Fernandez2016}.
Visual distractions are coined in the following terms: ``sleepiness'', ``drowsiness'', ``fatigue'', and ``inattention''.
And, they usually depend on facial landmarks detection and tracking.
Manual distractions are mainly concerned with driver's activities other than safe driving (i.e. reaching behind, adjusting hair and makeup, or eating and drinking).
In this kind of distraction, authors often tend to depend heavily on hand tracking and driving posture estimation.
In this paper, we focus only on ``manual'' distractions where a driver is distracted by texting or using cell phone, calling, eating or drinking, reaching behind, fiddling with the radio, adjusting hair and makeup, or talking to a passenger.

Recently, the field witnessed a dramatic increase in computational power (thanks to massive parallelization in modern GPUs) and available big data for Deep Learning training. That enabled more research on advanced/deep network architectures. One example is the proliferation of deep learning-based solutions utilizing convolutional neural networks to solve computer vision problems--
 which achieved unprecedented performance
 \cite{krizhevsky2012imagenet,dosovitskiy2016discriminative,russakovsky2015imagenet}.
 That improvement is a byproduct of learning feature maps (as in, \cite{krizhevsky2012imagenet,erhan2014scalable}) rather than hand-crafting them in traditional computer vision practices \cite{oliva2001modeling}.
 
 In this paper, we detect and identify distraction using Deep Learning. RGB images are obtained from a camera mounted above the dashboard. We train and benchmark multiple convolutional neural network architectures.
 We use pre-trained networks in a ``transfer learning''-style where networks are pre-trained on the Imagenet dataset \cite{russakovsky2015imagenet} (that is, 1.2 million images and 1000 classes). This initializes the network with a generic set of features and allows it to recognize a variety of objects with high precision. Then, we re-train the latent fully connected layers to recognize the distraction postures (instead of Imagenet classes). During training the convolutional layers are usually either kept frozen or updated with a miniscule learning rate.
 In such way, we transfer the networks knowledge about 1,000 different objects (i.e. the broad domain) to our more specific domain of 10 driving distraction postures.

We present a real-time system for driver distraction identification that uses a learnable weighted ensemble of Convolutional Neural Networks (CNNs), a new method for skin segmentation, a challenging distracted driver's dataset (see figure \ref{fig:data_samples}) on which we evaluate our proposed solution, and an annotation tool \cite{Abouelnaga2017} for action labeling that can be used to extend our dataset.


\section{Literature Review}

The work in the distracted driver detection field over the past seven years could be clustered into four groups: multiple independent cell-phone usage detection publications, Laboratory of Intelligent and Safe Automobiles in University of California San Diego (UCSD) datasets and publications, Southeast University Distracted Driver dataset and affiliated publications, and recently, StateFarm's Distracted Driver Kaggle competition.

\subsection{Cell Phone Usage Detection}

\cite{Berri2014} presents an SVM-based model that detects the use of mobile phone while driving (i.e. distracted driving).
Their dataset consists of frontal image view of a driver's face.
They also make pre-made assumptions about hand and face locations in the picture.
\cite{Artan2009} presents another SVM-based classification method to detect cell phone usage.
However, their dataset is collected from transportation imaging cameras that are deployed in highways and traffic lights which is, indeed, more competitive.
\cite{craye2015driver} uses AdaBoost classifier and Hidden Markov Models to classify a Kinect's RGB-D data.
Their solution depends on indoor-produced data.
They sit on a chair and a mimic a certain distraction (i.e. talking on the phone).
This setup misses two essential points: the lighting conditions and the distance between a Kinect and the driver.
In real-life applications, a driver is exposed to a variety of lighting conditions (i.e. sunlight and shadow).
\cite{Zhang2011} suggests using a Hidden Conditional Random Fields (HCRF) model to detect cell phone usage.
Their model operates face, mouth, and hand features of images obtained from a camera mounted above the dashboard.
\cite{HoangNganLe2016} devised a Faster-RCNN model to detect driver's cell-phone usage and ``hands on the wheel''.
Their model is mainly geared towards face/hand segmentation.
They train their Faster-RCNN on the dataset proposed in \cite{Das2015} (that we also use in this paper).
Their proposed solution runs at a 0.06, and 0.09 frames per second for cell-phone usage, and ``hands on the wheel'' detection.
\cite{Seshadri2015} tackles the problem of cell phone usage detection.
Their approach doesn't hold any static assumptions though (i.e. in which region of the image a face is expected to be found).
They use a Supervised Descent Method (SDM) to localize the face landmarks, and then, extract two bounding boxes to the left and the right side of the face.
They train a classifier on each of the two regions to detect cell phone usage: right hand, left hand, or no usage.
Using a histogram of gradients (HOG) and an AdaBoost classifier, they achieve a 93.9\% classification accuracy and operate in a near real-time speed (7.5 frames per second).

\subsection{UCSD's Laboratory of Intelligent and Safe Automobiles Work}


\cite{Martin2014} presents an vision-based analysis framework that recognizes in-vehicle activities using two Kinect cameras that provide frontal and back views of the driver.
Their approach provides ``hands on the wheel'' information (i.e. left hand only, both hands, no hands), and uses these information to detect three types of distractions: adjusting the radio, operating the gear, and adjusting the mirrors.
\cite{ohn2013driver} presents a fusion of classifiers where the image is to be segmented into three regions: wheel, gear, and instrument panel (i.e. radio).
It proposes a classifier for each segment to detect existence of hands in those regions.
The hand information (i.e. output of the classifiers) is passed to an ``activity classifier'' that infers the actual activity (i.e. adjusting the radio, operating the gear).
\cite{Ohn-bar2014a} extends existing research to include eye cues to previously existing head and hands cues.
However, it still considers three types of distractions: ``wheel region interaction with two hands on the wheel, gear region activity, and instrument cluster region activity''.
\cite{Ohn-bar2013} presents a region-based classification approach.
It detects hands presence in certain pre-defined regions in an image.
A model is learned for each region separately.
All regions are later joined using a second-stage classifier.

\subsection{Southeast University Distracted Driver Dataset}


\cite{zhao2012recognitionRF} designs a more inclusive distracted driving dataset with a side view of the driver and more activities: grasping the steering wheel, operating the shift lever, eating a cake and talking on a cellular phone.
It introduces a contourlet transform for feature extraction, and then, evaluates the performance of different classifiers: Random Forests (RF), $k$-Nearest Neighbors (KNN), and Multilayer Perceptron (MLP).
The random forests achieved the highest classification accuracy of 90.5\%.
\cite{zhao2012recognitionMLP} showed that using a multiwavelet transform improves the accuracy of Multilayer Perceptron classifier to 90.61\% (previously 37.06\% in \cite{zhao2012recognitionRF}).
\cite{Zhao2011} showed that using a Support Vector Machine (SVM) with an intersection kernel, followed by Radial Basis Function (RBF) kernel, achieved the highest accuracies of 92.81\% and 94.25\%, respectively (in comparison with \cite{zhao2012recognitionRF} and \cite{zhao2012recognitionMLP}).
After testing against other classification methods, they concluded that an SVM with intersection kernel offers the best real-time quality (67 frames per second) and better classification performance.
\cite{zhao2013recognition} improves the Multilayer Perceptron classifier using combined features of Pyramid Histogram of Oriented Gradients (PHOG) and spatial scale feature extractors.
Their Multilayer Perceptron achieves a 94.75\% classification accuracy.
\cite{Yan2014} utilizes Motion History Images (HMI) to make use of the data's temporality.
Pyramid Histogram of Gradients (PHOG) is applied to the motion history images.
A Random Forrest trains on the extracted features and yields a 96.56\% accuracy.
\cite{Yan2016DrivingDistraction} presents a convolutional neural network solution that achieves a 99.78\% classification accuracy.
They train their network in a 2-step process. First, they use pre-trained sparse filters as the parameters of the first convolutional layer.
Second, they fine-tune the network on the actual dataset.
Their accuracy is measured against the 4-classes of the Southeast dataset.

\subsection{StateFarm's Dataset}

StateFarm's Distracted Driver Detection competition on Kaggle was the first publicly available dataset for posture classification.
In the competition, StateFarm defined ten postures to be detected: safe driving, texting using right hand, talking on the phone using right hand, texting using left hand, talking on the phone using left hand, operating the radio, drinking, reaching behind, doing hair and makeup, and talking to passenger.
Our work, in this paper, is mainly inspired by StateFarm's Distracted Driver's competition.
While the usage of StateFarm's dataset is limited to the purposes of the competition \cite{Sultan2016}, we designed a similar dataset that follows the same postures.


\section{Dataset Design}

Creating a new dataset was essential to the completion of this work.
The available alternatives to our dataset are: StateFarm and Southeast University (SEU) datasets.
StateFarm's dataset is to be used for their Kaggle competition purposes only (as per their regulations) \cite{Sultan2016}.
As for Southeast University (SEU) dataset, it presents only four distraction postures.
And, after multiple attempts to obtain it, we figured out that the authors do not make it publicly available.
All the papers (\cite{Yan2016,Yan2016DrivingDistraction,Yan2014,zhao2013recognition,zhao2012recognitionMLP,Zhao2011,zhao2012recognitionRF}) that benchmarked against the dataset are affiliated with either Southeast University, Xi'an Jiaotong-Liverpool University, or Liverpool University, and they have at least one shared author.
With that being said, the collected ``distracted driver'' dataset is the first publicly available (obtainable after signing a license agreement) for driving posture estimation research.
Our dataset is is publicly available subject to signing our agreement form from \cite{AUC_dataset}. The dataset introduced in this work is an extended and cleaned-up version of our dataset presented in \cite{DistractionDetection_2018_NIPS}. 

\subsection{Camera Setup}
Our dataset collection setup has a single camera with a fixed perspective, and the data collection was conducted on two phases. In each phase, a different camera is used. In one phase we use the ASUS ZenFone smartphone (Model ZD551KL) rear camera \cite{ASUSZenFone}, and in the other phase we used the DS325 Sony DepthSense camera \cite{depthsenseCamera}. The latter camera provides depth information, but we only record the RGB images. Collecting data from different cameras adds an extra dimension of diversity to our dataset, and we demonstrate the feasibility of effective distraction detection by relying on RGB cameras which are widely available and low cost.

The data was collected in a video format, and then, cut into individual images, $1080 \times 1920$ or $640 \times 480$ each.
The camera's are fixed using an arm strap to the car roof handle on top of the front passenger's seat.
In our use case, this setup proved to be very flexible as we needed to collect data in different vehicles.

\subsection{Labeling}

In order to label the collected videos, we designed a simple multi-platform action annotation tool using modern web technologies: Electron, AngularJS, and Javascript.
The annotation tool is open-source and publicly available at \cite{Abouelnaga2017}.

\subsection{Statistics}

We had 44 participants from 7 different countries: Egypt (37), Germany (2), USA (1), Canada (1), Uganda (1), Palestine (1), and Morocco (1). Out of all participants, 29 were males and 15 were females. Some drivers participated in more than one recording session with different time of day, driving conditions, and wearing different clothes.
Videos were shot in 5 different cars: Proton Gen2, Mitsubishi Lancer, Nissan Sunny, KIA Carens, and a prototyping car. We extracted 14,478 frames distributed over the following classes: Safe Driving (2,986), Phone Right (1,256), Phone Left (1,320), Text Right (1,718), Text Left (1,124), Adjusting Radio (1,123), Drinking (1,076), Hair or Makeup (1,044), Reaching Behind (1,034), and Talking to Passenger (1,797). The sampling is done manually by inspecting the video files with eye and giving a distraction label for each frame. The transitional actions between each consecutive distraction types are manually removed. Figure \ref{fig:data_samples} shows samples for the ten classes in our dataset.


\section{Proposed Method}

\begin{figure*}[ht]
  \centering
  \includegraphics[width=\textwidth]{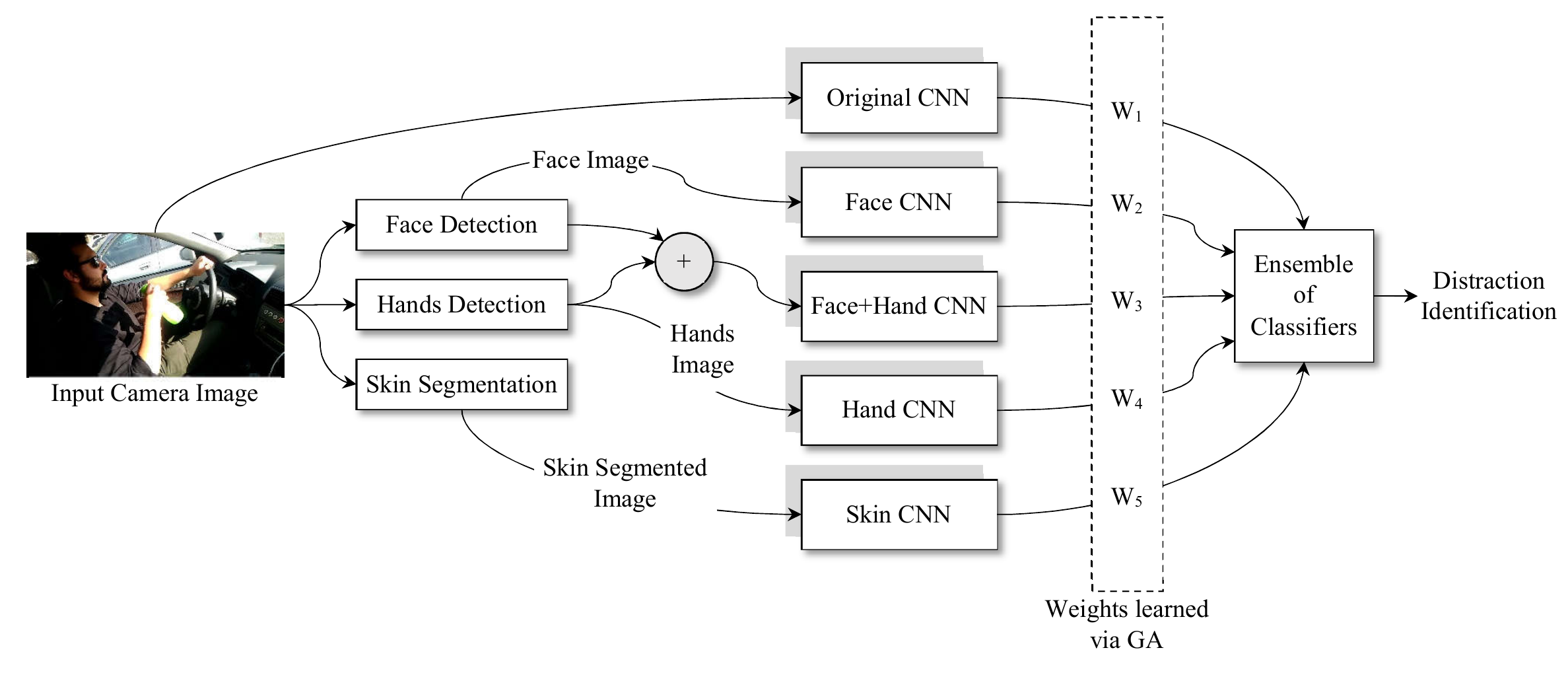}
  \caption{An overview of our proposed solution.
  A face detector, a hand detector, and a skin segmenter are run against each frame. For each output image (i.e. Skin, Face, Hands), an AlexNet and an InceptionV3 networks are trained (i.e. resulting in 10 neural networks: 5 AlexNet and 5 InceptionV3). The overall class distribution is determined by the weighted sum of all softmax layers. The weights are learned using a genetic algorithm.}
  \label{fig:system_overview}
\end{figure*}

Our proposed solution consists of a genetically-weighted ensemble of convolutional neural networks.
The convolutional neural networks are trained on raw images, skin-segmented images, face images, hands images, and ``face+hands'' images. On those five images sources, we train and benchmark an AlexNet network \cite{krizhevsky2012imagenet}, an InceptionV3 network \cite{szegedy2016rethinking}, a Resnet network having 50 layers \cite{he2016deep}, and a VGG-16 network \cite{{simonyan2014very}}. We fine-tune a pre-trained ImageNet model (i.e. transfer learning) for these networks. Then, we evaluate a weighted sum of all networks' outputs yielding the final class distribution using a genetic algorithm. The system overview is shown in figure \ref{fig:system_overview}.

\subsection{Skin Segmentation}

Skin segmentation is a challenging problem to solve, mainly due to the different lighting  conditions happening during driving. We use a Multivariate Gaussian Naive Bayes classifier to develop a pixel-wise skin segmentation model. Our model is similar to \cite{Phung2003} except that we do not use a histogram as a Likelihood function. Instead, we fit the training data into Gaussian distributions to formulate the Likelihood functions. The posterior probability is evaluated as in \ref{eq:bayesian_classifier1}.

\newcommand*{\Skin}{\,\text{Skin}}
\newcommand*{\NonSkin}{\,\text{non-skin}}
\newcommand*{\MuSkin}{\mu_{\Skin}}
\newcommand*{\MuNonSkin}{\mu_{\NonSkin}}
\newcommand*{\SigmaSkin}{\Sigma_{\Skin}}
\newcommand*{\SigmaNonSkin}{\Sigma_{\NonSkin}}
\newcommand*{\Prob}{\mathbb{P}}

\begin{equation} \label{eq:bayesian_classifier1}
    \Prob(skin | x) = \frac{\Prob(x | skin) \cdot \Prob(skin)}{\Prob(x)}, \,\,\,\,\, \Prob(x \mid skin) \sim \mathcal{N}(\MuSkin,\,\SigmaSkin)
\end{equation}

Note that $\Prob(skin) = \Prob(non\text{-}skin) = 0.5$ (i.e. we don't make any assumptions about existence of skin pixels in the image). 

We trained our model using the UCI Skin Segmentation dataset \cite{bhatt2010skin}. The database contains RGB colors that are labeled for the skin and non-skin classes, It is generated using skin textures from face images of diversity of age, gender, and race people. It contains a total of 245,057 color samples; out of which 50,859 is the skin samples and 194,198 is non-skin samples. Two Gaussian distributions (Likelihoods) are constructed for the skin and the non-skin classes by estimating $\mu_{skin}$, $\Sigma_{skin}$, $\mu_{non-skin}$, and $\Sigma_{non-skin}$ from the training data. For deployment phase, each pixel $x$ in the input image, is fed to the model as in Equation \ref{eq:bayesian_classifier2}. And then, a probability heat map of skin in the image can be constructed. We classify a pixel to a "skin" if $\text{Model}(x) > 0.5$.
Then, we cluster the skin pixels into objects and remove those with a small number of pixels. Because neither faces nor hands skin blobs are expected to have small number of pixels.

\begin{equation} \label{eq:bayesian_classifier2}
\begin{split}
    \text{Model}(x) & = \frac{\Prob(skin \mid x)}{\Prob(skin \mid x) + \Prob(non\text{-}skin \mid x)}\\
     & = \frac{\Prob(x \mid skin)}{\Prob(x \mid skin) + \Prob(x \mid non\text{-}skin)}
\end{split}
\end{equation}

\begin{figure*}[h]
    \centering
    \includegraphics[width=0.4\textwidth]{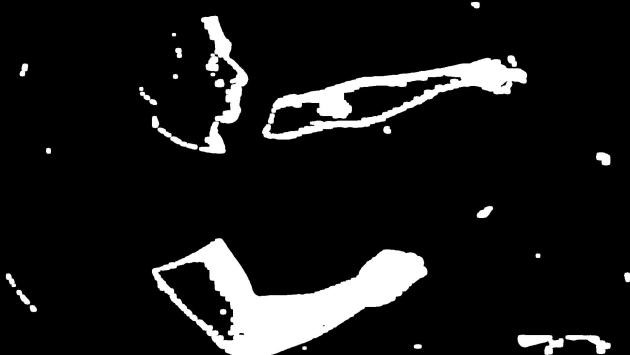}%
    \vspace{0.1cm}
    \includegraphics[width=0.4\textwidth]{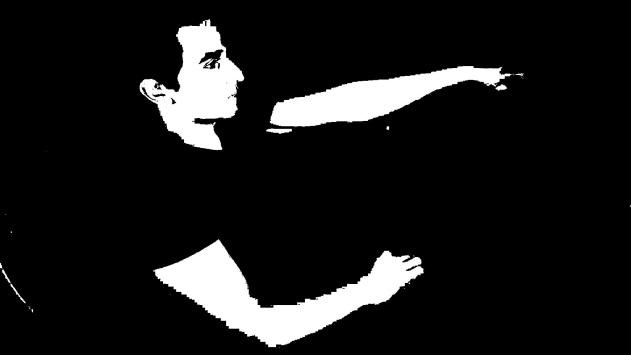}
    \caption{Left and right images show a sample for skin segmentation result without and with training the pixels spatial information respectively}
    \label{fig:skin_beforeAndAfter}
\end{figure*}

One key disadvantage of such method is that it's very sensitive to image illumination conditions;
Hence, incorporating pixel location can improve the skin classification accuracy.
One way is to pass the pixel location to the input feature vector. 
However, to the best of our knowledge, there is no available dataset to train and evaluate such method.
Besides, annotating a new dataset is costly.
Therefore, we adopt an active learning-based approach to supervise the training.
The above classifier (without pixel spatial information) is ran against all training images to generate skin masks.
Generated masks are manually inspected to cherry-pick samples with high skin segmentation accuracy.
Those pixels are used as new training data of the proposed skin segmentation classifier, such that the feature vector includes pixels spatial information (X and Y-coordinates within the images) in addition to the color information (Red, Green, and Blue color components). Figure \ref{fig:skin_beforeAndAfter} shows a sample skin-segmented image with (right) and without (left) pixel spatial information. We notice an accuracy improvement after considering the pixels spatial information. More test data results are presented in the Experiments section \ref{sec:experiments}.

\subsection{Face \& Hands Detection}
We trained the model presented in \cite{LiHaoxiangandLinZheandShenXiaohuiandBrandtJonathanandHua2015} on the Annotated Facial Landmarks in the Wild (AFLW) face dataset \cite{tugraz:icg:lrs:koestinger11b}. It was sensitive to distance from the camera; faces that were close to the camera were not easily detected. We found that the pre-trained model (presented in \cite{Farfade2015}) produced better results on our dataset. Given that we did not have any hand labelled face bounding boxes, we couldn't formally compare the two models. But it was obvious that \cite{Farfade2015} gives a better detection accuracy based on inspecting the results manually. However, face misdetections are noticed in several examples, mainly because the detector is not trained to handle non-frontal faces. 

\begin{figure*}[ht]
  \centering
  \includegraphics[width=0.7\textwidth]{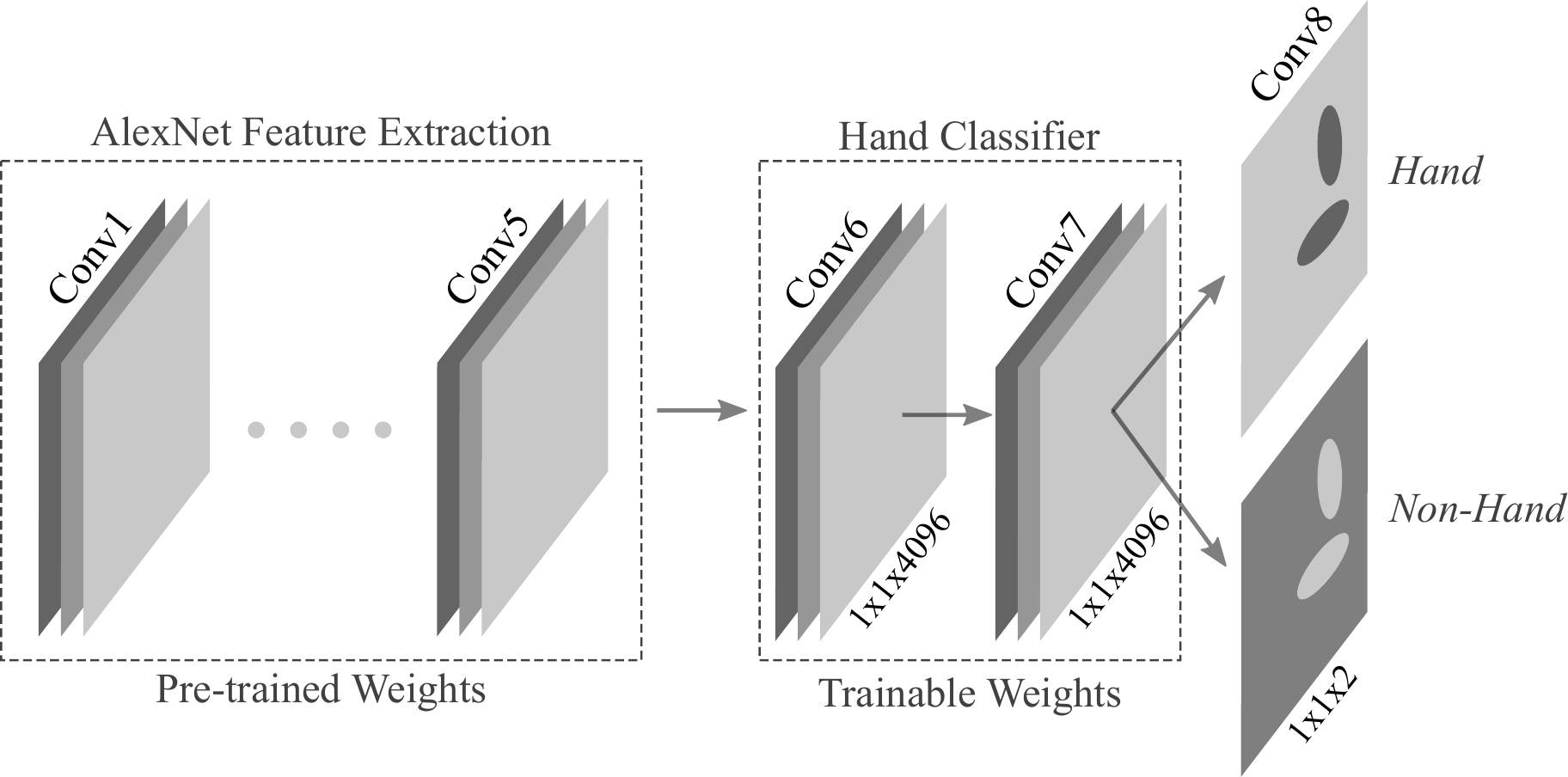}
  \caption{
    A modified version of AlexNet for hands localization ("Conv" stands for a convolutional layer).
    An AlexNet is trained on "hands" and "non-hands" images.
    Fully connected layers are replaced with convolutional layers with the same weights of the fully connected layers.
  }
  \label{fig:hand_detector}
\end{figure*}

As for hands detection, we used the pre-trained model presented in \cite{Bambach_2015_ICCV} with modifications. Their trained model was a binary class AlexNet that classifies hands/non-hands for different proposal windows. We transferred the weights of the fully connected layers (i.e. fc6, fc7 and fc8) into convolutional layers such that each neuron in the fully connected layer was transferred into a feature map layer with a 1-pixel kernel size.
Our proposed architecture, shown in figure \ref{fig:hand_detector}, accepts variant size inputs and produces variant-size outputs.
The last convolutional layer has a depth of 2 (i.e. the binary classes), and for each pixel the summation of the two depths is one as in \ref{eq:hand_detect}, where $W$ and $H$ are the output's width and height respectively.

\begin{equation}
\label{eq:hand_detect}
\begin{split}
{Conv8}_{x,y,0} + {Conv8}_{x,y,1} = 1, \quad & 0 \le x \le W \\
                                                  & 0 \le y \le H
\end{split}
\end{equation}

\subsection{Convolutional Neural Network}

For distracted driver posture classification, we trained and benchmarked different neural networks architectures: an AlexNet \cite{krizhevsky2012imagenet}, an InceptionV3 \cite{szegedy2016rethinking}, a Resnet network having 50 layers \cite{he2016deep}, and a VGG-16 network \cite{simonyan2014very}.
Each network is trained on 5 different image sources (i.e. raw, skin, face, hands and face+hands images).

We trained our AlexNet models from scratch. We didn't use a pre-trained model.
As for InceptionV3, we performed transfer learning.
We fine-tuned a pre-trained model \cite{TensorflowSlim} on the distraction postures.
We removed the ``logits'' fully connected layer, and replaced it with 10-neuron fully connected layer (i.e. corresponding to 10 driving postures).
For all of our models, we used a gradient descent optimizer with an initial learning rate of $10^{-2}$.
The learning rate decays linearly in each epoch with a step of $\frac{10^{-2} - 10^{-4}}{\text{Epochs}}$.
We trained the networks for 30 epochs. In each epoch, we divide the training dataset into mini-batches of 50 images each.

\subsection{GA-based Ensemble of Classifiers}

Each classifier produces a class probability vector (i.e. output of the ``softmax'' layer), $C_1 \,...\,\, C_N$, such that $C_i  \in \mathbb{R}^{10}$ is a vector having 10 probabilities (for 10 distraction classes) and $N$ is the number of classifiers.
In a majority voting system as in Equation \ref{eq:average_ensambling}, it is assumed that all experts (i.e. classifiers) can equally contribute to a better decision by taking the unweighted sum of all classifier outputs.
\begin{equation} \label{eq:average_ensambling}
C_{\text{Majority}} = \frac{1}{N} \sum_{i}^{N} C_i
\end{equation}
However, that is not usually a valid assumption.
In a weighted voting system as in Equation \ref{eq:weighted_ensambling}, we assume that classifiers do not contribute equally to the ensemble and that some classifiers might yield higher accuracy than others.
Therefore, there is a need to estimate the weights of each classifier's contribution to the ensemble. \cite{rokach2010ensemble} presents a variety of methods to estimate the weights.
We opted to use a genetic algorithm (i.e. a search-based method).

\begin{equation} \label{eq:weighted_ensambling}
C_{\text{Weighted}} = \frac{1}{\sum_{i}^{N} w_i} \sum_{i}^{N} w_i \cdot C_i
\end{equation}

In our genetic algorithm, a chromosome consists of $N$ genes that correspond to the weights $w_1 \,\, ... \,\, w_N$.
Our fitness function evaluates the Negative Log Likelihood (NLL) loss over a 50\% random sample of the population.
This helps prevent overfitting.
Our population consists of 50 individual.
In each iteration, we retain the top 20\% of the population and use them as parents.
Then, we randomly select 10\% of the remaining 80\% of the population as parents.
In other words, we have 30\% of the population as parents.
Now, we randomly mutate 5\% of the selected parents.
Finally, we cross-over random pairs of the parents to produce children until we have a full population (i.e. with 50 individuals).
We ran the above procedure for only 5 iterations in order to avoid over-fitting.
We selected the chromosome with the highest fitness score (test against all data points-- not 50\%).

\section{Experiments}
\label{sec:experiments}

We divided our dataset into 75\% for training and 25\% held out as test data using random sampling.
Then, we ran the skin segmentation, face and hand detectors on the entire dataset. We tested different deep networks against our test dataset and obtained the results reported in table \ref{table:accuracy_test_data_split_randmly}.
We notice that both AlexNet and InceptionV3 achieve best accuracies when trained on the original images.
However, the accuracy doesn't majorly change in both architectures when switching from the original images to skin segmented images.
Hands seem to have more weight in posture recognition than the face.
``Face + Hands'' images produce slightly lower accuracy than the hands images alone, yet still higher than the face images alone.
That happens due to face/hand detector failures.
For example, if a hand is not found, we pass a face image to a ``face + hands'' classifier.
This doesn't happen in individual cases of hand-only or face-only classifier because if the hand/face detection fails, we pass the original image to the hand/face classifier as a fallback mechanism.
With better hand/face detectors, the ``face+hands'' networks are expected to produce higher accuracies than the ``hands'' networks.
The confusion matrix of our genetically weighted ensemble of classifiers on this randomly selected test data is shown in table \ref{table:confusion_matrix_test_data_split_randmly}, and the results are shown in table \ref{table:accuracy_test_data_split_randmly}.

\begin{table}[ht]
\def\arraystretch{1.5} 
\centering
\arrayrulecolor{black} 
\caption{Distracted driver posture classification results on randomly selected test data}
\label{table:accuracy_test_data_split_randmly}
\begin{tabular}{|c|c|c|c|}
\hline
Model                            & Source               & Loss (NLL)            & Accuracy (\%)  \\ \hline
\multirow{5}{*}{AlexNet}         & \textbf{Original}    & \textbf{0.3909} & \textbf{93.65} \\ \cline{2-4}
                                 & Skin Segmented       & 0.3468          & 93.62           \\ \cline{2-4}
                                 & Face                 & 1.0516          & 84.28          \\ \cline{2-4}
                                 & Hands                & 0.6186          & 89.52          \\ \cline{2-4}
                                 & Face + Hands         & 0.8298          & 86.68          \\ \hline
\multirow{5}{*}{InceptionV3}     & \textbf{Original}    & \textbf{0.2654} & \textbf{95.17} \\ \cline{2-4}
                                 & Skin Segmented       & 0.2903          & 94.66          \\ \cline{2-4}
                                 & Face                 & 0.6096          & 88.82          \\ \cline{2-4}
                                 & Hands                & 0.4546          & 91.62          \\ \cline{2-4}
                                 & Face + Hands         & 0.4495          & 90.88          \\ \hline
\multicolumn{2}{|c|}{AlexNet} & 0.2727 & 94.29 \\ \hline
\multicolumn{2}{|c|}{\textbf{Majority Voting Ensemble}} & \textbf{0.1661} & \textbf{95.77} \\ \hline
\multicolumn{2}{|c|}{\textbf{GA-Weighted Ensemble}}     & \textbf{0.1575} & \textbf{95.98} \\ \hline
\end{tabular}
\end{table}


\begin{table}[ht]
\def\arraystretch{1.5} 
\small
\caption{Confusion matrix of our genetically weighted ensemble of classifiers on randomly selected test data}
\label{table:confusion_matrix_test_data_split_randmly}
\centering
\newcommand\items{10}   
\arrayrulecolor{white} 
\noindent\begin{tabular}{cc*{\items}{|E}|}
\multicolumn{1}{c}{\centering} &\multicolumn{1}{c}{} &\multicolumn{\items}{c}{Predicted} \\ \hhline{~*\items{|-}|}
\multicolumn{1}{c}{} & 
\multicolumn{1}{c}{} & 
\multicolumn{1}{c}{C0} & 
\multicolumn{1}{c}{C1} & 
\multicolumn{1}{c}{C2} & 
\multicolumn{1}{c}{C3} & 
\multicolumn{1}{c}{C4} & 
\multicolumn{1}{c}{C5} & 
\multicolumn{1}{c}{C6} & 
\multicolumn{1}{c}{C7} & 
\multicolumn{1}{c}{C8} & 
\multicolumn{1}{c}{C9} \\ \hhline{~*\items{|-}|}
\multirow{\items}{*}{\rotatebox{90}{Actual}} 
&C0     &95.34  &0      &0.33   &0.65   &0.11   &0.43   &0.43   &0.87   &0.11   &1.74   \\ \hhline{~*\items{|-}|}
&C1     &0.31   &96.63  &1.23   &0.31   &0.92   &0      &0.31   &0      &0.31   &0      \\ \hhline{~*\items{|-}|}
&C2     &0.29   &3.23   &96.48  &0      &0      &0      &0      &0      &0      &0      \\ \hhline{~*\items{|-}|}
&C3     &2.02   &0.61   &0      &96.15  &0.81   &0      &0.20   &0      &0      &0.20   \\ \hhline{~*\items{|-}|}
&C4     &0      &0.33   &0      &4.90   &94.77  &0      &0      &0      &0      &0      \\ \hhline{~*\items{|-}|}
&C5     &4.26   &0      &0      &0.33   &0      &95.08  &0      &0      &0      &0.33   \\ \hhline{~*\items{|-}|}
&C6     &0.74   &0      &0      &0.25   &0      &0.74   &98.01  &0.25   &0      &0      \\ \hhline{~*\items{|-}|}
&C7     &3.65   &0      &0      &0      &0      &0      &0      &95.35  &0      &1.00   \\ \hhline{~*\items{|-}|}
&C8     &3.79   &0      &0      &0      &0      &0      &1.38   &0.34   &92.76  &1.72   \\ \hhline{~*\items{|-}|}
&C9     &1.40   &0      &0      &0      &0      &0      &0.47   &0.31   &0.16   &97.67  \\ \hhline{~*\items{|-}|}
\end{tabular}
\end{table}

\begin{figure}[ht]
  \centering
  \includegraphics[width=1.0\textwidth]{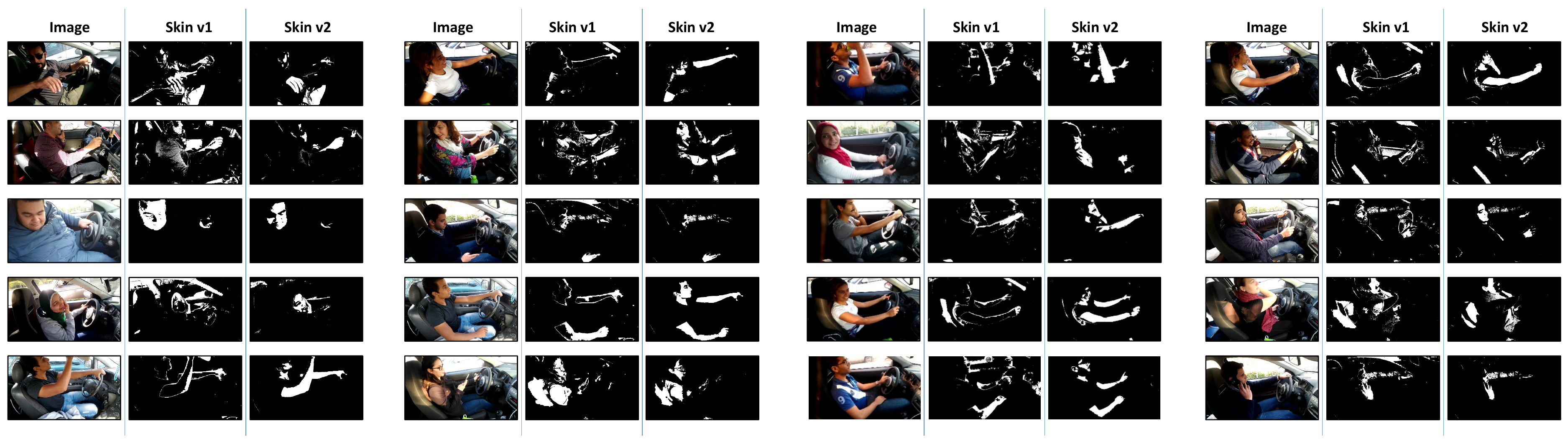}
  \caption{
  Our skin segmentation algorithm (v1 \& v2) is run against sample test images. "Skin v2" is trained on pixel spatial information, while "Skin v1" is only trained on pixel color information.}
  \label{fig:skins}
\end{figure}

In figure \ref{fig:skins}, we show sample results of our skin segmentation algorithm. Test data are never seen by the skin segmentation component during training. In the "Skin v2" column, the classifier is trained on pixels' spatial information. In the "Skin v1" column, it is only trained on pixels color information. In addition to such qualitative evaluation for our skin detection method, in table \ref{table:skin_spatial_quantitative_results} we report a quantitative comparison of distraction classification accuracy by training on skin images generated with and without using pixels' spatial information. An AlexNet is trained and tested on skin segmented images.

\begin{table}[h]
\def\arraystretch{1.5} 
\centering
\caption{Distraction classification results by training (AlexNet) and testing on skin segmented images}
\label{table:skin_spatial_quantitative_results}
\arrayrulecolor{black} 
\begin{tabular}{|c|c|}
\hline
Network     & Accuracy {[}\%{]} \\
\hline
No pixels' spatial info                                            & 93.14          \\\hline
Spatial info learned through active learning with 50\% of data     & 93.53          \\\hline
Spatial info learned through active learning with 100\% of data    & 93.62          \\
\hline
\end{tabular}
\end{table}

\begin{figure}[b]
  \centering
  \includegraphics[width=0.75\textwidth]{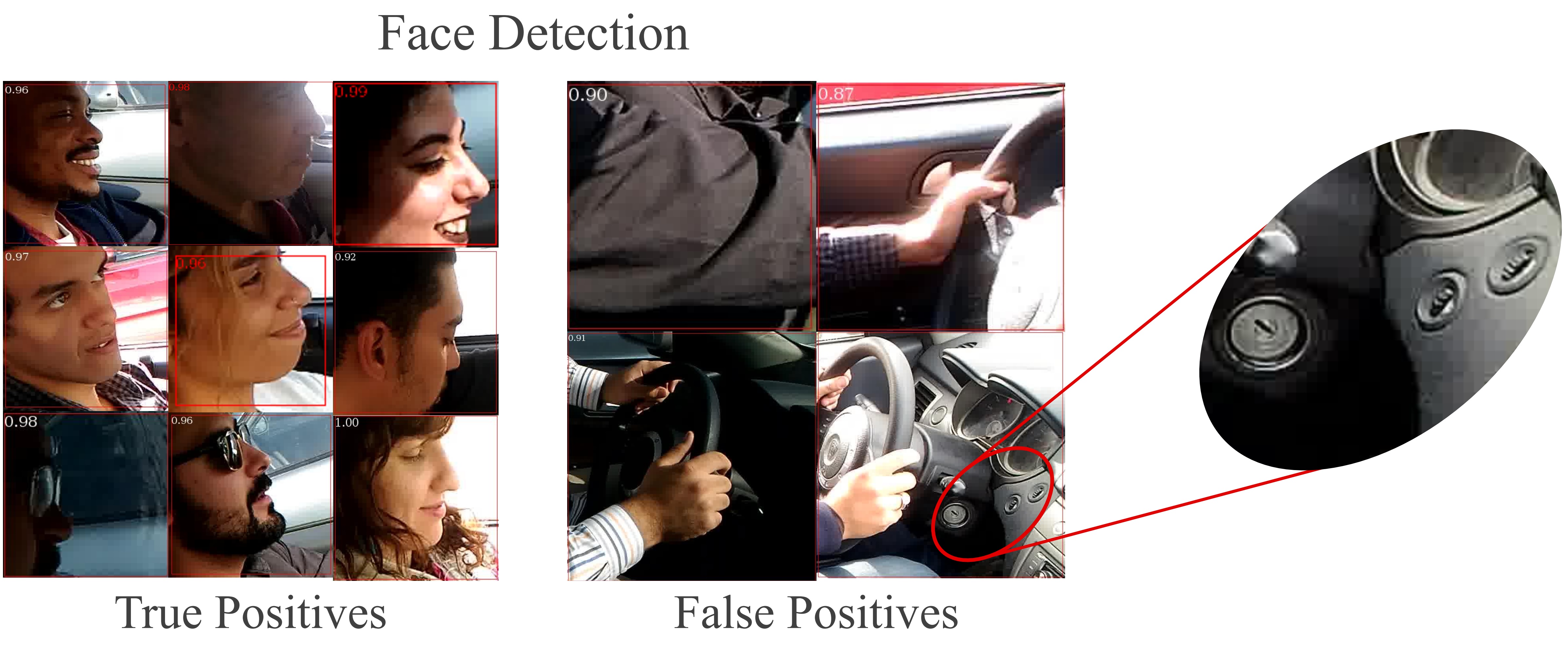}
  \caption{Non-frontal faces detection is challenging. A correctly detected face and a wrongly detected one by \cite{Farfade2015}. Air-conditioning openings are wrongly detected as a face.}
  \label{fig:face_detector_results}
\end{figure}

Table \ref{table:skin_spatial_quantitative_results} first row result is for a model trained on images generated without spatial information. In the third row, active learning is used to supervise learning pixels' spatial information with pixels generated from 8,662 cherry-picked skin segmented images. 100 pixels are randomly sampled from each image to provide an annotated training dataset of 866,200 pixels. A third model is trained on only 50\% of these pixels and the result is reported in the second row.
The spatial information is proven to improve the skin detection accuracy, while the overall method performance remains unsatisfactory and needs better generalization against the varying lighting conditions happening during driving. Given that, and in addition to several misdetections noted from the face (as in figure \ref{fig:face_detector_results}) and hand detectors, we conclude that further work is needed to improve the accuracy of detection of these visual elements (skin, face, and hands). At the moment they just add to the system complexity without improving the overall distraction identification accuracy.

The train/test data split presented in table \ref{table:accuracy_test_data_split_randmly} causes a high correlation between training and testing data.
Thus, it makes the distraction detection a much easier problem. This split is also followed in \cite{baheti2018detection}. To present a more realistic setup, we adopt a more challenging strategy. We separate the data based on driving sessions rather than at random. In other words, test data split contains different drivers, cars, times of day, lighting conditions, driving conditions and so on. Out of 44 drivers in our dataset, the training contains 38 drivers (12,555 samples), and the test data contains 6 drivers (1,923 samples).

\begin{figure}[ht]
  \centering
    \includegraphics[width=0.8\textwidth]{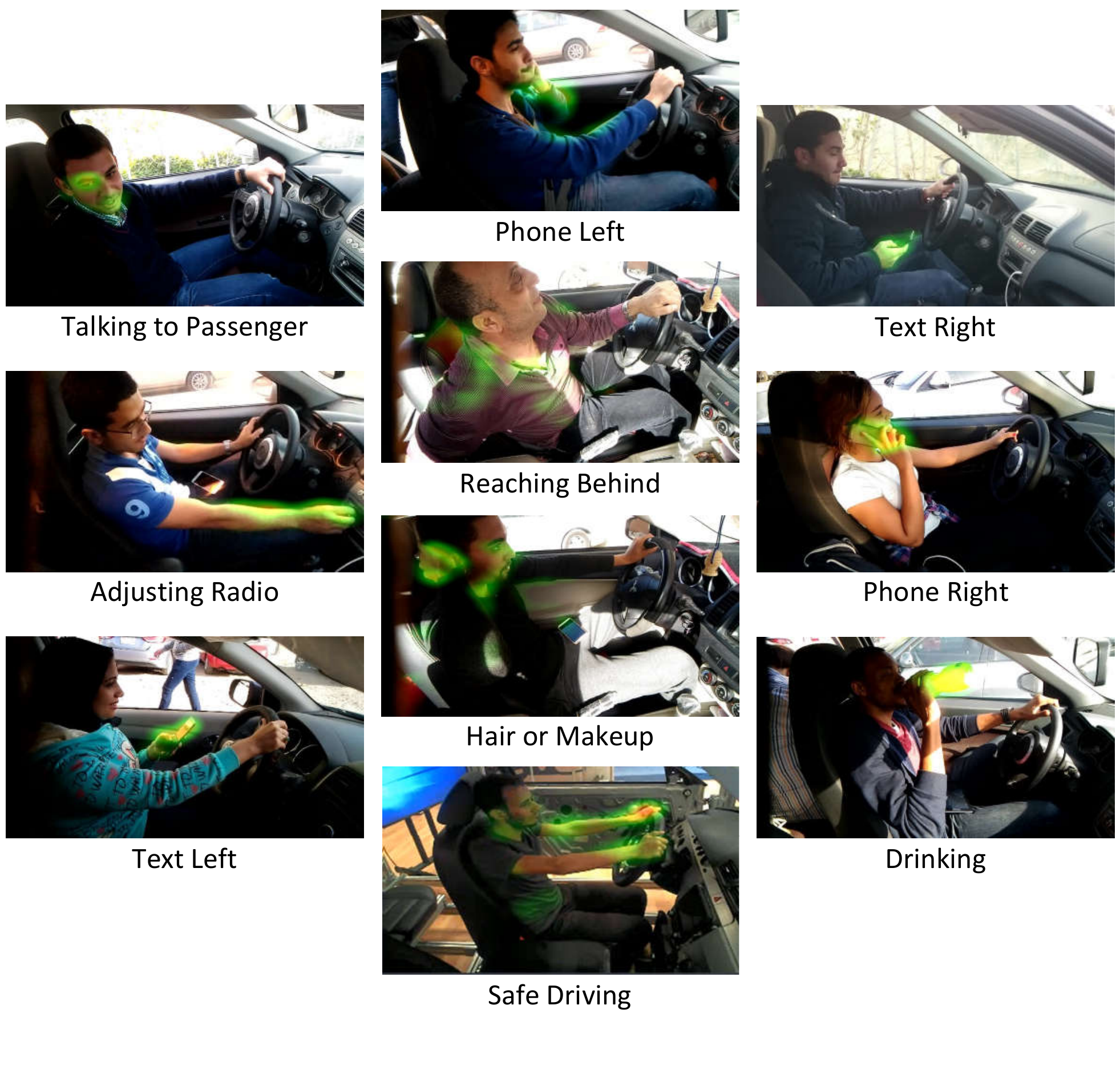}
    \caption{Saliency map demonstrates that our network makes its decision based on relevant features/regions in the input camera images.}
    \label{fig:saliency_maps}
\end{figure}

The results of our genetically weighted ensemble of classifiers on the new split-by-driver test data is shown in table \ref{table:accuracy_test_data_split_by_driver}. The confusion matrix is shown in table \ref{table:confusion_matrix_test_data_split_by_driver}. InceptionV3 achieves the best distraction identification accuracy of 90\%. Figure \ref{fig:saliency_maps} presents saliency maps generated for sample test images. The gradient of the winning distraction class is calculated and applied to input image. The gradient highlights the salient regions-- i.e. the ones causing the most change and contributing to the network's decision. The drawn salient regions clearly show that our network is making its decision based on relevant features in the input camera images (i.e. hands, faces, neck, radio, driving wheel and so on).

\begin{table}[ht]
\def\arraystretch{1.5} 
\centering
\caption{Distracted driver posture classification results on test data with unique drivers}
\label{table:accuracy_test_data_split_by_driver}
\arrayrulecolor{black} 
\begin{tabular}{|c|c|c|}
\hline
Network     & Loss (NLL) & Accuracy {[}\%{]} \\
\hline
VGG-16      & 1.246614   & 76.131          \\\hline
Resnet50    & 0.661483   & 81.695          \\\hline
InceptionV3 & 0.640018   & 90.068          \\
\hline
\end{tabular}
\end{table}

\begin{table}[ht]
\def\arraystretch{1.5} 
\small
\centering
\caption{Confusion matrix of InceptionV3 network on test data with unique drivers}
\label{table:confusion_matrix_test_data_split_by_driver}
\newcommand\items{10}   
\arrayrulecolor{white}   
\noindent\begin{tabular}{cc*{\items}{|E}|}
\multicolumn{1}{c}{\centering} &\multicolumn{1}{c}{} &\multicolumn{\items}{c}{Predicted} \\ \hhline{~*\items{|-}|}
\multicolumn{1}{c}{} & 
\multicolumn{1}{c}{} & 
\multicolumn{1}{c}{C0} & 
\multicolumn{1}{c}{C1} & 
\multicolumn{1}{c}{C2} & 
\multicolumn{1}{c}{C3} & 
\multicolumn{1}{c}{C4} & 
\multicolumn{1}{c}{C5} & 
\multicolumn{1}{c}{C6} & 
\multicolumn{1}{c}{C7} & 
\multicolumn{1}{c}{C8} & 
\multicolumn{1}{c}{C9} \\ \hhline{~*\items{|-}|}
\multirow{\items}{*}{\rotatebox{90}{Actual}}
&C0     &87.57  &10.98  &0      &0.29   &0      &0    &0.29   &0.86   &0      &0     \\ \hhline{~*\items{|-}|}
&C1     &0      &96.71  &0      &1.88   &0      &0    &0      &1.41   &0      &0     \\ \hhline{~*\items{|-}|}
&C2     &0      &0      &97.42  &0      &0      &0    &0      &0.52   &0.52   &1.55  \\ \hhline{~*\items{|-}|}
&C3     &8.33   &0      &0      &91.67  &0      &0    &0      &0      &0      &0     \\ \hhline{~*\items{|-}|}
&C4     &11.76  &0      &0      &0      &87.06  &0    &0      &0      &1.18   &0     \\ \hhline{~*\items{|-}|}
&C5     &0      &0      &0      &0      &0      &100  &0      &0      &0      &0     \\ \hhline{~*\items{|-}|}
&C6     &0.67   &0      &0      &0      &0      &0    &81.82  &10.49  &6.99   &0     \\ \hhline{~*\items{|-}|}
&C7     &0      &0      &0      &0      &0      &0    &0      &100    &0      &0     \\ \hhline{~*\items{|-}|}
&C8     &2.74   &0      &0.68   &0      &3.42   &0    &0      &7.53   &84.93  &0.68  \\ \hhline{~*\items{|-}|}
&C9     &1.38   &2.75   &0      &0      &0      &0    &0      &19.27  &0      &76.6  \\ \hhline{~*\items{|-}|}
\end{tabular}
\end{table}

In a static image, a driver would appear in a "safe driving" posture, while contextually he/she might be distracted by doing some other activity. Making the distraction posture identification decision based only on the current camera input image restricts the system from coping with such a problem. In this experiment, to classify a  camera frame $F_t$ at timestamp $t$, the network gives a decision $C_t$. After that, a final smoothed decision is deduced as the average of all the preceding network decisions over the $M$ seconds before $t$ such that $C_{smoothed_t} = \sum_{i=t-M}^{t} C_i $.

\begin{figure}[ht]
\centering
\def\axisdefaultwidth{0.75\textwidth}
\def\axisdefaultheight{3cm}
\input{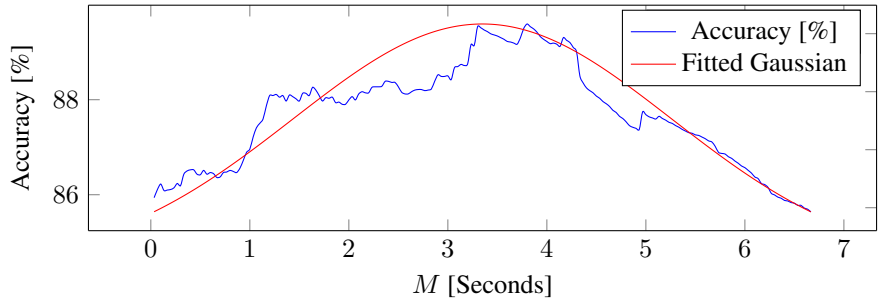}
\caption{System accuracy evaluated at different values of history intervals $M$. At $M=1$, current camera input image is solely used for distraction classification; No context is used. For $M > 1$, we average predictions of $M$ frames to obtain distraction in the current frame.
Up to a certain threshold, the more context incorporated in the prediction, the more accurate and robust the system is.
After that threshold, contextual information in the past is no longer relevant.
Best accuracy is achieved at $M = 3.35$ seconds-- which is the average of the fitted Gaussian curve.
}
\label{fig:averaging_analysis}
\end{figure}

In figure \ref{fig:averaging_analysis}, we evaluate the system accuracy at different values of $M$. $M=1$ means that distraction posture identification decision is based only on the current camera input image. A bell-curve-looking graph is obtained; It indicates that accuracy increases when more images from the past are considered. However, this happens until a threshold after which the past becomes irrelevant; thus, decreasing accuracy. The optimal value for $M$ is found to 3.35 seconds, which is the mean of a fitted Gaussian distribution to the results.

\subsection{Analysis}

In table \ref{table:confusion_matrix_test_data_split_randmly}, we notice that the most confusing posture is the ``safe driving''.
This is due to the lack of temporal context in static images.
In a static image, a driver would appear in a ``safe driving'' posture.
However, contextually, he/she was distracted by doing some other activity.
``Text Left'' is mostly confused for ``Talk Left'' and vice versa.
Same applies to ``Text Right'' and ``Talk Right''.
``Adjust Radio'' is mainly confused for a ``safe driving'' posture.
That is due to lack of the previously mentioned temporal context.
Apart from safe driving, ``Hair \& Makeup'' is confused for talking to passenger.
That is because, in most cases, when drivers did their hair/makeup on the left side of their face, they needed to tilt their face slightly right (while looking at the frontal mirror).
Thus, the network thought the person was talking to passenger.
``Reach Behind'' was confused for both talking to passenger and drinking.
That makes sense as people tend to naturally look towards the camera while reaching behind.
As for the drinking confusion, it is due to right-arm movement from the steering wheel to the back seat.
A still image in the middle of that move could be easily mistaken for a drinking posture.
``Drink'' and ``Talk to Passenger'' postures were not easily confused with other postures as 98\% and 97.67\% of their images were correctly classified.

\subsection{Real-time System}
The number of parameters in the deep models we adopted and benchmarked in \ref{table:accuracy_test_data_split_randmly} is huge for a real-time system. Our version of VGG16 \cite{simonyan2014very}, AlexNet \cite{krizhevsky2012imagenet}, Resnet50 \cite{he2016deep}, and InceptionV3 \cite{szegedy2016rethinking} has 134.3M, 58.3M, 23.6M, and 21.8M parameters, respectively. \cite{baheti2018detection} adopts a real-time model that reduces the number of parameters in VGG16 to 15M, which is a marked improvement, but is still a large number for a real-time system (in a self-driving car, for example). Hence we propose a version of NasNetMobile \cite{zoph2017learning} that reduces the number of parameters further to only 4.3M. An ensemble of two NasNetMobile models \cite{zoph2017learning} (Original and Skin-segmented networks) produce a satisfactory classification accuracy of 84.64\%. Meanwhile, it still maintains a real-time performance on a CPU-based system. 


For NasNetMobile models fusion, we use a Multilayer Perceptron (MLP) to learn a class-based weighted ensemble from data.
The advantage of our MLP-based fusion over the weighted voting system is that it does not assume that some classifier is better in absolute terms.
An MLP learns a more sophisticated dependencies; one classifier can be more accurate in discriminating a set of distraction classes, while another can be more accurate on another set of distraction classes. This also can happen under specific status of the ensemble resultant softmax class probability distributions. In our adopted MLP-based fusion, the vectors $C_i \in \mathbf{R}^{10}, 
\text{where} \,\, i = 1, ..., N$, are concatenated and passed to an MLP. Note that, each $C_i$ is a vector having 10 probabilities (corresponding to 10 distraction classes). Our MLP loss function evaluates the Negative Log Likelihood (NLL) loss over the training data in addition to a regularization term to prevent over-fitting.


\section{Conclusion}
\label{sec:conclusion}

Distracted driving is a major problem leading to a striking number of accidents worldwide.
Its detection is an important system component in semi-autonomous cars.
In this paper, we presented a robust vision-based system that recognizes distracted driving postures.
We collected a novel publicly-available distracted driver dataset that we used to develop and test our system.
Our best model utilizes a genetically weighted ensemble of convolutional neural networks to achieve a 90\% classification accuracy. We aim to provide a baseline performance for future research to benchmark against.
We also showed that a simpler model (only using AlexNet) could operate in real-time and still maintain a satisfactory classification accuracy.
Face, hands, and skin detection is proved to improve classification accuracy in our ensemble.
However, in a real-time setting, their performance overhead is much higher than their contribution.

In a future work, we need to devise a better face, hands, and skin detector.
We would need to manually label hand and face proposals and use them to train a Fast-RCNN (or, any other object detector) to localize both faces and hands in one shot and evaluate it against our existing CNN-based localization method.


\acknowledgments{%
This work was supported by the German Research Foundation (DFG) and the Technical University of Munich within the Open Access Publishing Funding Programme.%
}


\bibliography{library.bib}  

\end{document}